\documentclass[journal,twoside]{IEEEtran}
\usepackage{microtype}
\usepackage{graphicx}
\usepackage{subfigure}
\usepackage{booktabs} 
\usepackage{amssymb,amsmath}
\usepackage{pifont}
\usepackage{color}
\usepackage[english]{babel}
\usepackage{array}
\usepackage{bm}
\usepackage{multirow}
\usepackage{enumitem}
\usepackage{pifont}
\usepackage{tikz}
\usepackage{amssymb}
\usepackage{enumitem}
\usepackage[english]{babel}
\usepackage{array}
\usepackage{bm}
\usepackage{multirow}
\usepackage{enumitem}
\usepackage{pifont}
\usepackage{graphicx}
\newcommand{\eat}[1]{}
\usepackage{flushend}

\newcommand{\todo}[1]{\textcolor[rgb]{0.00,0.00,0.00}{#1}}
\newcommand{\todoww}[1]{\textcolor[rgb]{0.00,0.00,0.00}{#1}}
\newcommand{\todoqw}[1]{\textcolor[rgb]{0.00,0.00,0.00}{#1}}

\newcommand{\todoly}[2]{\textcolor[rgb]{0.00,0.00,0.00}{#1}}

\def\placeholder{NACIM}

\begin{document}

\title{Device-Circuit-Architecture Co-Exploration for Computing-in-Memory Neural Accelerators}

\author{Weiwen Jiang,
        Qiuwen Lou,
        Zheyu Yan,
        Lei Yang,
        Jingtong Hu,//
        Xiaobo Sharon Hu~\IEEEmembership{Fellow,~IEEE},
        and Yiyu Shi~\IEEEmembership{Senior Member,~IEEE}
\thanks{
   W. Jiang, Q. Lou, Z. Yan, L. Yang, X. Hu and Y. Shi are with the Department of Computer Science and Engineering, University of Notre Dame, Notre Dame, IN 46556 (e-mail: wjiang2@nd.edu; yshi4@nd.edu).
}
\thanks{J. Hu is with the Department of Electrical and Computer Engineering, University of Pittsburgh, Pittsburgh, PA 15261.
}
\thanks{This work is partially supported by the National Science Foundation under Grants CCF-1919167, CCF-1820537 and CNS-1822099. 
}
}

\maketitle

\begin{abstract}
Co-exploration of neural architectures and hardware design is promising due to its capability to simultaneously optimize network accuracy and hardware efficiency.
However, state-of-the-art neural architecture search algorithms for the co-exploration are dedicated for the conventional von-Neumann computing architecture, whose performance is heavily limited by the well-known memory wall.
In this paper, we are the first to bring the computing-in-memory architecture, which can easily transcend the memory wall, to interplay with the neural architecture search, aiming to find the most efficient neural architectures with high network accuracy and maximized hardware efficiency.
Such a novel combination makes opportunities to boost performance, but also brings a bunch of challenges: The optimization space spans across multiple design layers from device type and circuit topology to neural architecture; and the presence of device variation may drastically degrade the neural network performance.
To address these challenges, we propose a cross-layer exploration framework, namely \placeholder, which jointly explores device, circuit and architecture design space and takes device variation into consideration to find the most robust neural architectures, coupled with the most efficient hardware design.
Experimental results demonstrate that \placeholder~can find the robust neural network with 0.45\% accuracy loss in the presence of device variation, compared with a 76.44\% loss from the state-of-the-art NAS without consideration of variation; in addition, \placeholder~achieves an energy efficiency up to 16.3 TOPs/W, 3.17$\times$ higher than the state-of-the-art NAS.


\end{abstract}

\begin{IEEEkeywords}
Hardware/Software Co-Design; Computing-in-Memory Architecture; Neural Architecture Search; Neural Network Accelerator.
\end{IEEEkeywords}

\setlength{\textfloatsep}{3pt}
\setlength{\floatsep}{3pt}
\setlength{\dbltextfloatsep}{3pt}

\section{Introduction} \label{sec:Intro}

After deep neural network achieved great success, we are now witnessing the process of Artificial Intelligence (AI) democratization, which involves various machine learning tasks (e.g., image classification, video segmentation, speech recognition) \cite{krizhevsky2012imagenet,redmon2018yolov3}, tremendous applications (e.g., automotive vehicle, robot, health care) \cite{gao2018object,xu2019whole} and different hardware platforms (e.g., CPUs, GPUs, FPGAs, ASICs) \cite{zhang2018thundervolt,zhang2018fate,jiang2019achieving,xu2018resource,jiang2018heterogeneous}.
\todo{One of the most important questions} in the AI democratization era is: \textit{Given a dataset with a specified machine learning task, how to efficiently identify the best neural network architecture and hardware design, such that the network accuracy and hardware efficiency can be maximized simultaneously.}

To solve this problem, Neural Architecture Search (NAS) \cite{zoph2016neural, zoph2017learning, real2017large, liu2017hierarchical, nekrasov2019architecture, liu2018darts} has been proposed to liberate human labor in the design of neural architectures by automatically identifying their hyperparameters.
However, such an approach does not take hardware into consideration, which may easily lead the identified architecture to be useless due to the violation of the required hardware specifications.
\todoqw{To address this deficiency, hardware-aware NAS} \todo{\cite{tan2018mnasnet, cai2018efficient, wu2018fbnet, cai2018proxylessnas, dai2019chamnet,stamoulis2019single}} has been proposed, in which the hardware specifications are considered during the search process.
\todoww{To further improve hardware efficiency, co-exploration of neural architectures and hardware design is proposed in \cite{jiang2019accuracy,hao2019fpga,jiang2019hardware}, which proves that the Pareto frontiers between network accuracy and hardware efficiency can be further pushed forward by opening the hardware design space.}
However, all the works are based on the conventional von-Neumann architecture (e.g., mobile platform or FPGAs), leading the memory accesses inevitably becoming the performance bottleneck due to the well-known memory wall.

Computing-in-memory (CiM) has been proved to be able to effectively transcend such a memory wall \cite{wong2018inmemory}, and has been considered to be a promising candidate for neural network computations due to the incomparable architectural benefits.
{(i)} CiM architecture can benefit from the fixed memory access pattern within neural network computation \cite{sze2017tutorial} to execute operations in place. {(ii)} Emerging devices (e.g., ReRAM, STT-RAM) can be efficiently leveraged in the in-memory computing architecture \cite{shafiee2016isaac} to provide high performance and energy efficiency.
In \cite{biswas2018convram,kang2018in_memory}, MOSFET based in-memory processing has been employed for neural network computation, and the improvement in terms of energy and delay are observed compared with the conventional von-Neumann architectures.
\todoww{Research works \cite{shafiee2016isaac,chi2016prime} leverage emerging devices based in-memory computing scheme to construct crossbar architectures that can perform the matrix multiplication in analog domain, which further optimizes the computation metrics such as area, energy, and delay.}

\todo{Most of the existing works on CiM neural accelerator design simply map classic neural networks (e.g., LeNet, AlexNet) to the CiM platform to evaluate their design and compare against other counterparts.} However, without the optimization on neural architectures, these reported metrics (i.e., accuracy, latency, energy, etc.) may be far from the optimal. 
\todoww{In this work, we bring the CiM neural accelerator design to interplay with the neural architecture search, aiming to automatically identify the best device, circuit, and neural architecture coupled with the maximized network accuracy and hardware efficiency.} 
To the best of our knowledge, this is the first work to carry out the device-circuit-architecture co-exploration for CiM neural accelerators.


The novel device-circuit-architecture co-exploration brings opportunities to boost performance; however, it also incurs many new challenges.
First of all, unlike the conventional von-Neumann architecture based neural architecture co-exploration \cite{jiang2019hardware}, the design space of CiM-based neural accelerator spans across multiple layers from device type, circuit topology to neural architecture.
Second, limited by the computing capacity of each device cell, quantization is essential to improve the hardware efficiency \cite{xu2018scaling,xu2018quantization,zhang2019compact}; as such, quantization has to be automatically determined during the search process. 
\todoww{Third, in addition to the optimization goals of hardware efficiency used in the existing co-exploration framework for mobile platform and FPGAs, CiM has extra objectives, such as minimizing area, maximizing lifetime, etc.}
Last but not least, emerging devices commonly have non-ideal behaviors (known as device variation); that is,
if we directly map the trained DNN models to the architecture  without considering the device variation, a dramatic accuracy loss will be observed, rendering the architecture useless.

This paper proposes a device-circuit-architecture co-exploration framework, namely \placeholder, to automatically identify the best CiM neural accelerators, including the device type, circuit topology, and neural architecture hyperparamters.
\placeholder~framework will iteratively conduct explorations based on a reward function, which is suitable for reinforcement learning approaches or evolutionary algorithms.
By configuring the parameters of the framework, designers can customize the optimization goals in terms of their demands.
Furthermore, we have considered the device variation in the framework. 
In the forward path of our training framework, we incorporate the variation in the computation, which is based on the device noise model \cite{zhao2017rram}.
Experimental results show that the proposed \placeholder~framework can find the robust neural network with only 0.45\% accuracy loss in the presence of device variation, compare with a 76.44\% loss from the state-of-the-art NAS without considering device variation.
In addition, \placeholder~can significantly push forward the Pareto frontier in terms of the tradeoff between accuracy and hardware efficiency, achieving up to 16.3 TOPs/W energy efficiency for a 3.17$\times$ improvement.


The main contributions of this work are listed as follows.

\begin{itemize}[noitemsep,topsep=0pt,parsep=0pt,partopsep=0pt]
  \item We formally define the optimization problem of identifying the best computing-in-memory (CiM) neural accelerator, whose design space spans across device type, circuit topology to neural architecture. To the best of our knowledge, this is the first work on optimizing CiM neural accelerators together with neural architecture search.
  \item We have proposed a novel device-circuit-architecture co-exploration framework, namely \placeholder, to simultaneously optimize network accuracy and hardware efficiency. The framework further optimizes the quantization to boost the hardware efficiency and considers the device variation to identify the robust neural architectures.
  \item We implement the \placeholder~framework using a reinforcement learning approach and evaluate it on the commonly used datasets. Experimental results demonstrate the efficacy of the proposed framework in identifying the robust neural architectures in terms of device variation and pushing forward the Pareto frontier between accuracy and efficiency.
\end{itemize}

The remainder of the paper is organized as follows. 
Section \ref{sec:pre} presents the background of both neural architecture search and computing-in-memory architectures. 
Section \ref{sec:pdef} demonstrates the search space of five layers, and formally defines the cross-layer optimization problem. 
The proposed novel cross-layer optimization framework is presented in Section \ref{sec:frame}.
Experimental results are shown in Section \ref{sec:exp}.
Finally, concluding remarks are given in Section \ref{sec:conclusion}.

\section{Background}\label{sec:pre}

\subsection{System-Level Overview}

\begin{figure}[t]
\begin{center}
\centerline{\includegraphics[width=3.3 in]{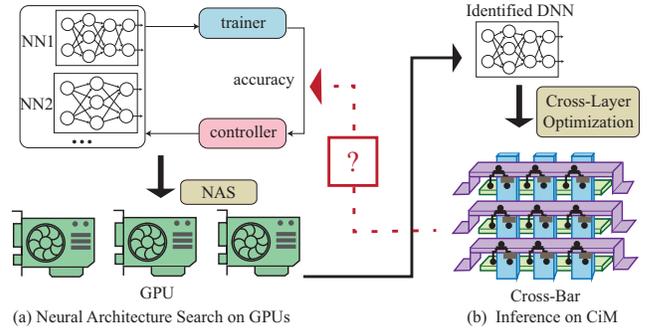}}
\vspace{-10pt}
\caption{An overview of neural architecture search phase and the accelerated inference phase: (a) we based on GPU to train child networks during the NAS, and (b) the identified neural network will be finally deployed to the target Computing-in-Memory (CiM) architecture to accelerate the inference.}
\label{overview}
\end{center}
\end{figure}

Figure \ref{overview} demonstrates the overview of extending the conventional framework of neural architecture search to optimize neural architectures for the  non-volatile devices based computing-in-memory architecture.
Specifically, the neural architecture search process is first performed on GPUs, which involves the training of new models from scratch to generate the reward.
After the search process is convergent, the identified neural network architecture will finally be deployed on the target computing-in-memory architecture.
However, as shown in Figure \ref{overview}, there is a missing link between the neural architecture search process and the computing-in-memory neural accelerator design.
We will introduce the neural architecture search and computing-in-memory platform in the following subsections.

\subsection{Neural Architecture Search}

Most recently, Neural Architecture Search (NAS) has been consistently achieving breakthroughs in different machine learning applications, such as image classifications \cite{zoph2016neural}, image segmentation \cite{liu2019auto}, video action recognition \cite{peng2019video}, etc.
NAS attracts large attentions mainly because it successfully releases human expertise and labor to identify high-accuracy neural architectures.

A typical NAS, such as that in \cite{zoph2016neural}, is composed of a controller and a trainer.
The controller will iteratively predict neural architecture parameters, called child network, and the trainer will train the child network from scratch on a held-out data set to obtain its accuracy.
Then, the accuracy will be feedback to update controller.
Finally, after the number of child networks predicted by the controller exceeds a predefined threshold, the search process will be terminated. 
Among all of the searched neural architectures, the one with the highest accuracy will be finally identified.

It has been demonstrated in existing works that the automatically searched neural architectures can achieve similar accuracy to the best human-invented architectures \cite{zoph2016neural,zoph2017learning}.
However, the identified architectures may have more complicated structures, which reduse their usefulness in real-world applications. For instance, it will result in excessive bandwidth requirement to perform secured inference.

\subsection{Computing-in-Memory}
In this paper, we consider the crossbar as the basic compute-in-memory engine. We discuss the devices used in this work, and the non-ideal behavior of the device. We also adopt NeuroSim, the framework we used to simulate crossbar computation.

\subsubsection{{Device and its variations}}
Non-volatile devices have been widely adopted in the crossbar computations.  When considering using the crossbar to perform inference, different device implementations lead to distinct energy, latency, etc. Here, we consider two factors (1) how many levels of precision the non-volatile device can be configured; (2) the non-ideal behavior of the devices. Both binary devices and multi-level devices are used in existing crossbar-based computation platforms. For the multi-level device, there are existing works with 4-bit (i.e., 16 levels) devices, with good distinction among different levels \cite{zhao2017rram}. Besides the multi-level devices, binary devices (STT-MRAM, etc.) are also considered in our implementation. Different kinds of devices may affect the on and off current for the crossbar computation, and ultimately impact delay, energy, etc. Different number of levels in these devices also requires different peripheral circuitries in the crossbar architecture, which is another design space we will consider in this work.

These emerging devices also suffer from various errors \cite{ben2018error}. When the circuitry is used for inference, device-to-device variations could be the dominant error source. The variation could be caused in the fabrication process and in the device programming phase. The other dominant sources of error come from noises. Among the noise sources, random telegraph noise (RTN)~\cite{ben2018error} in particular, is a main source of noise caused by electrons temporarily being trapped within the device which in turn changes the effective conductance of device. Other noise sources include thermal noise and shot noise. However, they typically are much smaller compared with RTN~\cite{ben2018error}. In this work, we model the device variation as a whole, and use a Gaussian distribution to represent the variation. The magnitude of the variation can be referred from \cite{zhao2017rram}, where the variations are from actual measurements.

\subsubsection{Crossbar Architecture}
Different crossbar based architectures are proposed \cite{shafiee2016isaac, chi2016prime}. We assume an ISAAC-like architecture \cite{shafiee2016isaac} in our simulation. The architecture is highly parallel with multiple tiles. Within each tile, there are multiple crossbar arrays. The computation here is performed in analog domain. However, ADC and DAC are used to convert the signal from and to the analog domain computation. We assume that all the weights can be mapped to the crossbar arrays. Therefore, no programming of the weights is needed in the computation. 

\subsubsection{{NeuroSim}}
DNN+NeuroSim \cite{peng2019dnn} is an integrated framework built for emulating the deep neural networks (DNN) inference performance or on-chip training performance on the hardware accelerator based on near-memory computing or in-memory computing architectures. Various device technologies are supported, including SRAM, emerging non-volatile memory (eNVM) based on resistance switching (e.g. RRAM, PCM, STT-MRAM), and ferroelectric FET (FeFET). SRAM is by nature 1-bit per cell, eNVMs and FeFET in this simulator can support either 1-bit or multi-bit per cell. NeuroSim \cite{chen2018neurosim} is a circuit-level macro model for benchmarking neuro-inspired architectures (including memory array, peripheral logic, and interconnect routing) in terms of circuit-level performance metrics, such as chip area, latency, dynamic energy and leakage power. With Pytorch and TensorFlow wrapper, DNN+ NeuroSim framework can support hierarchical organization from the device level (transistors from 130 nm down to 7 nm, eNVM and FeFET device properties) to the circuit level (periphery circuit modules such as analog-to-digital converters, ADCs), to chip level (tiles of processing-elements built up by multiple sub-arrays, and global interconnect and buffer) and then to the algorithm level (different convolutional neural network topologies), enabling instruction-accurate evaluation on the inference accuracy as well as the circuit-level performance metrics at the run-time of inference.

\section{Problem Definition}\label{sec:pdef}

\begin{figure}[t]
\begin{center}
\centerline{\includegraphics[width=3.5 in]{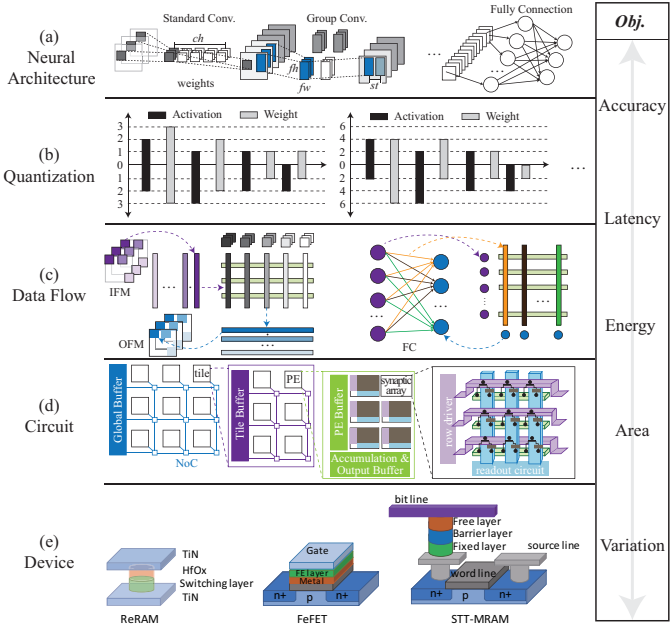}}
\vspace{-10pt}
\caption{Cross-layer optimization to identify the best neural architecture on computing-in-memory platform: (a) neural architecture; (b) 2 possible quantization for 4 layers; (c) data flow of generating output feature maps by using the input feature maps and weights; (d) layout of circuit; (e) different computing-in-memory devices.}
\label{crosslayer}
\end{center}
\end{figure}

Figure \ref{crosslayer} illustrates the cross-layer optimization from application to hardware.
\todoqw{Our ultimate goal is to implement the inference of a neural network on computing-in-memory (CiM) systems. Optimization decisions need to be made in five design layers, including (a) neural architecture search, (b) quantization determination, (c) data flow, (d) circuit design, and (e) device selection. In this section, we first introduce the detailed design options in all five design layers. Then, we discuss the search space derived from these design layers, and formally define the optimization problem.}

\subsection{\todo{Definitions of Cross-Layer CiM System}}
(a) Neural Architecture:
As shown in Figure \ref{crosslayer}, a neural architecture is composed of multiple layers, which is defined as $A=\langle L,para,acc\rangle$. 
It consists of a set of layers $L$.
The number of layers in the neural architecture is the size of set $L$, i.e., $|L|$.
A layer can be a convolutional layer, a fully connected layer, etc.
In order to automatically identify the neural architecture, we parameterize each layer to form a search space.
For the $i^{th}$ layer $l_i\in L$, set $para_i$ contains the predictable parameters, such as the number of filters and the filter size for convolution layer, and the number of neurons in the fully connected layer.
After we determined the parameters of all layers, we obtain a neural architecture, called \textit{child network}.
The accuracy of the child network is named $acc$, which can be obtained by training $A$ on a held-out dataset.
For illustration purpose, we use a linear chain of layers as an example.
However, the proposed technique is not limited to such structure and is applicable to more complicated structures, such as Directed Acyclic Graph (DAG), which can represent the residual connections.

(b) Quantization: For each layer of the neural architecture, we can apply different data precision for computation.
We define the quantization of a neural architecture $A=\langle L,para,acc\rangle$ as $Q(A)=\langle qa, qw \rangle$, where $qa$ and $qw$ represent the quantization for activation and weights, respectively.
For a layer $l_i \in L$, $qa_i=(M,N)$ indicates that we apply $M$ bits to represent the integer part and $N$ bits to represent the fraction part of the activation data; similarly, $qw_i=(P,Q)$ is defined for weights.
Figure \ref{overview} (b) illustrates two quantization instances for a 4-layer neural architecture, where the number above x-axis indicates the bit-width for integer part and the number below x-axis indicates the fraction part.

(c) Data Flow:
\todoqw{The data flow layer is the intermediate layer between software (neural architecture) and hardware (circuit and device). In terms of the pattern of data reuse, data flow can be classified into four categories: \textit{i)} weight stationary; \textit{ii)} output stationary; \textit{iii)} row stationary; and \textit{iv)} no local reuse. Taking weight stationary as an example, its basic idea is described as follows.}
First, for the convolution operation, the weights of a kernel are expanded and spread on the memory cells of cross-bar vertically; while for fully connection, the weights for each output neural are vertically spread on the cross-bar.
Second, the activation (i.e., IFM or input neural) is fed horizontally into the cross-bar.
Third, at each cycle, dot product is performed on the fed activation and the stationed weights to get the partial sums of outputs, and the accumulation operation is conducted on top of the previous obtained partial sums.
Figure \ref{crosslayer} (c) shows the above details for both convolution operation (left-hand side) and fully connection operation (right-hand side).




(d) Circuit:
Figure \ref{crosslayer} (d) shows the chip hierarchy. 
A chip is defined as $C=\langle T,PE,S,D \rangle$, which is composed of tile array $T$, PE array $PE$, and synaptic array $S$, and the device $D$.
The top-level of the chip is a network-on-chip (NoC) based $M\times N$ tile array, which is defined as $T=\langle M, N, buf, band \rangle$, where \textit{buf} is the size of the global buffer, and $band$ is the bandwidth of a link on NoC.
Similarly, a tile is composed of a $P\times Q$ PE array, which is defined as $PE=\langle P,Q,buf,band\rangle$; and a PE is composed of a $U\times V$ synaptic array, which is defined as $S=\langle U,V\rangle$.
In the synaptic array, each cell is a device, which is specified from a set of available devices defined as follows.


(e) Device: We will have different choices of devices to be employed in the circuit.
We define $DT=\langle T, bit, var \rangle$, where $T$ is a set of available devices (e.g., ReRAM, FeFET, STT-MRAM, as shown in Figure \ref{crosslayer} (e)).
For a specific device $t_i\in T$, say ReRAM, $bit_i=4$ indicates the applied ReRAM has the ability to store 4 bits in one cell; and $var_i$ refers to the variation function, which is based on the existing work (e.g., \cite{zhao2017rram} for ReRAM).
Kindly note that if the bit-width of a layer (in terms of $Q(C)$) is larger than $bit_i$, we adopt a shift-and-add circuitry at the peripheral, and we use multiple devices to represent the weights. Otherwise if the bit-width is less than $bit_i$, we employ one device to store the weights.
By leveraging the shift-and-add operation, we can achieve arbitrary the number of bits, which can well support the design space exploration when applying NAS to the crossbar. 


\begin{figure*}[t]
\begin{center}
\centerline{\includegraphics[width=6 in]{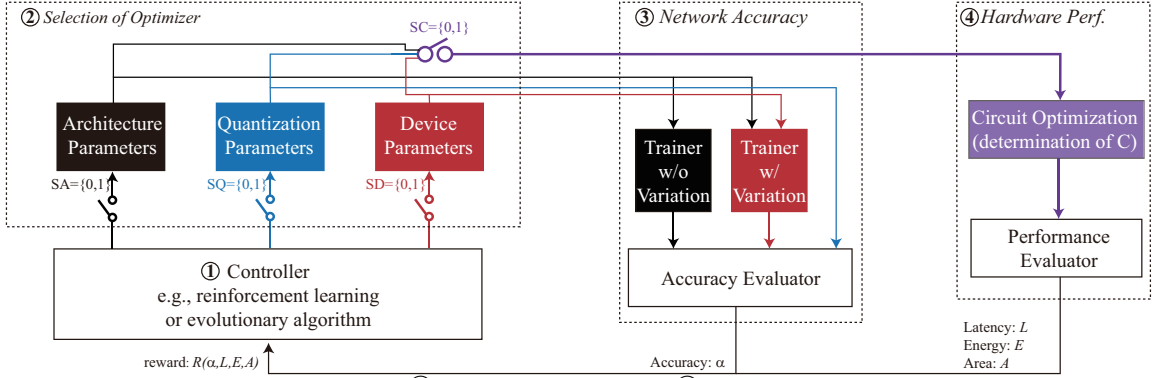}}
\vspace{-10pt}
\caption{Overview of the proposed \placeholder~framework: \raisebox{-1pt}{\large\ding{192}} a reward-based controller; \raisebox{-1pt}{\large\ding{193}} an optimizer selector for architecture A, quantization Q, device D, and circuit C; \raisebox{-1pt}{\large\ding{194}} an accuracy evaluator for identified neural architecture; \raisebox{-1pt}{\large\ding{195}} a hardware performance evaluator with the circuit optimization.}
\label{framework}
\end{center}
\end{figure*}

\subsection{\todo{Search Space and Problem Definition}}

\todoqw{\textit{Search Space:} The design spaces of all the layers form an integrated search space. Among the five design layers, the data flow design layer has the fewest options. Although there are different types of data flows in terms of the data reuse pattern, the weight-stationary data flow is commonly used for the CiM platform. In this work, we also apply weight-stationary data flow in the exploration. All the other design layers provide various design options. For the neural architecture layer, the size of the neural architecture can be adjusted to fit the hardware, which can be implemented by searching for the hyperparameters of the backbone neural architecture. For the quantization layer, different bit-widths for both integer and fraction parts can be employed for network layers. For the circuit layer, tile size, buffer size, and bandwidth should be determined. Finally, for the device layer, we have choices in different types of devices.}

\textit{Problem Statement:} Based on the definition of each layer, we formally define the problem solved in this work as follows: Given a dataset (e.g., CIFAR-10), a machine learning task (e.g., image classification), and a set of available devices $DT$, we are going to determine:
\begin{itemize}
\item $A$: the neural architecture for the machine learning task;
\item $Q$: the quantization of each layer in the architecture $A$;
\item $D$: the device in set $DT$ used for the chip design; 
\item $C$: the circuit design based on the selected device $D$; 
\end{itemize}
\textbf{Objective:} such that the inference accuracy of the machine learning task on the resultant circuit can be maximized, while the hardware efficiency (e.g., latency, energy efficiency, area, etc.) can be optimized.
Kindly note that since the above optimization problem has multiple objectives, we further propose a framework in the next section, which can support designers to specify the metrics to be optimized (e.g., \todoly{simultaneously maximizing accuracy, latency, and area---simultaneously maximizing accuracy, and minimizing latency and area}).

\section{Cross-Layer Exploration Framework}\label{sec:frame}


Figure \ref{framework} demonstrates the overview of the proposed Neural Architecture and Computing-in-Memory Architecture Co-Exploration Framework, named \placeholder, to solve the problem defined in Section \ref{sec:pdef}.
\placeholder~contains 4 components: 
\raisebox{-1pt}{\large\ding{192}} a controller
\raisebox{-1pt}{\large\ding{193}} an optimizer selector,
\raisebox{-1pt}{\large\ding{194}} \todo{a network accuracy evaluator}, 
\raisebox{-1pt}{\large\ding{195}} a hardware performance evaluator.

\raisebox{-1pt}{\large\ding{192}} \textbf{{Controller.}} The controller is a core component in
\placeholder~framework. 
\todo{It conducts optimizations on the neural architecture search and the CiM hardware design, where the optimizations can be implemented by different solvers, such as the reinforcement learning approach or evolutionary algorithm.}
Specifically, the controller predicts the hyperparameters of neural architecture, quantization, and device, according to the network accuracy and hardware performance from evaluators.
These metrics form a reward function for updating the controller. 
The reward function is formulated as follows.
\begin{equation}\label{equ:reward}
    R(\alpha, \beta) = \beta\times \alpha + (1-\beta)\times f(Lat, Eng, Area),
\end{equation}
where $\alpha$ is the prediction accuracy, $\beta$ is a scaling parameter, and $Lat,Eng,Area$ represent three hardware performance metrics: latency, energy, area.
\todo{These performance metrics will be determined by the design parameters related to architecture, quantization, and circuit.
We will introduce how to obtain these metrics later, in \raisebox{-1pt} {\large\ding{194}} {Accuracy Evaluator} and \raisebox{-1pt} {\large\ding{195}} {Performance Evaluator}.}
The merge function $f$ can either be a simple weighted sum or other more advanced functions defined by the user.
In Sec. \ref{sec:exp}, we adopt weighted sum for this function.

In terms of the reward, the controller will predict hyperparamters, which can be implemented by different techniques, such as the reinforcement learning approach or evolutionary algorithm. 
In this work, we employ the reinforcement learning method in the controller.
\todoww{Like the existing reinforcement learning based on neural architecture search \cite{zoph2016neural,jiang2019accuracy}, a recurrent neural network (RNN) is implemented in the controller for the prediction of the hyperparameters of a child network.
In our framework, as shown in Figure \ref{framework}, there are three kinds of hyperparameters: architecture parameters (e.g., the number of channels for each layer), the quantization parameters (e.g., the bit-width of integer and fraction part), and circuit/device parameters (e.g., which device to be used).
All possible combinations of these parameters form the state space in reinforcement learning.
In each iteration, the RNN predicts a set of hyperparameters, which is the action of reinforcement learning.
At the end of an iteration, we update the RNN network for better prediction in terms of the reward.
The update procedure is the interaction of the controller with the environment, which is modeled as a Markov Decision Process (MDP) for optimization. 
Specifically, the Monte Carlo policy gradient algorithm \cite{williams1992simple} is employed:}

\begin{equation}
    \nabla J(\theta) = \frac{1}{m}\sum_{k=1}^{m}\sum_{t=1}^{T}\gamma^{T-t}\nabla_{\theta}\log \pi_{\theta} (a_{t}|a_{(t-1):1})(R_{k}-b)
\end{equation}
where $m$ is the batch size and $T$ is the total number of steps in each episode. The rewards are discounted at every step by
an exponential factor $\gamma$ and the baseline $b$ is the average exponential moving of the reward.

\raisebox{-1pt}{\large\ding{193}} \textbf{{Optimizer Selector.}}
The optimizer selector will determine the flow in \placeholder~framework.
As shown in Figure \ref{framework} \raisebox{-1pt}{\large\ding{193}}, there are four switches $SA$, $SQ$, $SD$, $SC$ corresponding to four determination variables of neural architecture $A$, quantization $Q$, device $D$, and circuit $C$.
In terms of the status of switches, \placeholder~can perform different functions as listed in the following:
\begin{itemize}[noitemsep,topsep=0pt,parsep=0pt,partopsep=0pt]
  \item $SA=1,SQ=0,SD=0,SC=0$\\ 
  In the first case, \placeholder~performs the conventional neural architecture search, like \cite{zoph2016neural}, which aims to maximize accuracy without considering the hardware efficiency. 
  \item $SA=1,SQ=1,SD=0,SC=0$\\
  In the second case, \placeholder~considers the quantization during the neural architecture search, like \cite{Lu2019Neural}, which will simultaneously determine the neural architecture and the quantization for each network layer.
  \item $SA=1,SQ=1,SD=1,SC=0$\\
  In the third case, \placeholder~additionally involves the devices in the search process where the device variation will be considered to guarantee no accuracy loss after implementing the identified network on the target hardware.
  \item $SA=0,SQ=1,SD=0,SC=1$\\
  In the fourth case, \placeholder~further explores the circuit design space for circuit optimization together with quantization in terms of a given architecture and device. 
\end{itemize}

In this work, in order to conduct cross-layer optimization, we first set the switch combinations to the third case (called ``hardware perturbation aware NAS'', abbreviating as ``\todo{ptbNAS}''), such that we can identify neural architectures with high accuracy on the target devices with variation.
Second, we apply the fourth switch combination (called ``hardware resource aware NAS'', abbreviating as ``rNAS'') to further explore the circuit optimization to involve the hardware performance into consideration.
The details for \todo{ptbNAS} and rNAS will be introduced in the following two evaluators.


\raisebox{-1pt}{\large\ding{194}} \textbf{{Accuracy Evaluator.}}
The accuracy evaluator is the key component to execute \todo{ptbNAS}.
In the conventional neural architecture search based on the mobile or FPGA platforms, there is no need to consider hardware perturbation; however, when it comes to computing-in-memory based platform, the fundamental devices will have variations in their characteristics (i.e., device non-idealities), which in turn will affect the accuracy. 
As a result, if we do not consider the variation during training, as shown in the left component in Figure \ref{framework} \raisebox{-1pt}{\large\ding{194}}, there will be a dramatic accuracy loss when the identified architecture is deployed to the circuit.

The crossbar architecture is assumed for inference in this paper. However, the non-ideal behavior of the device in the inference stage may significantly decrease the application level accuracy \cite{zhang2019resilience}, which is a main concern when using the emerging devices in the crossbar architecture. In this work, we propose to use a modified training method to alleviate the impact of non-ideal behavior of the device and circuit, as shown in the right component of Figure \ref{framework} \raisebox{-1pt}{\large\ding{194}}.
When considering device variation in the training phase, the training typically requires a much longer time \cite{zhang2019resilience} than a conventional training method. \todo{As a result, leveraging existing methods will dramatically increase the search time. 
This will further extend the NAS search process, leading the framework inefficient.}
In this paper, \todo{we propose a method} to reduce the effects of device variation in a more efficient way.
Specifically, we propose a novel training method that invloves the device variation in the training procedure.
The method is composed of two steps: First, we use Monte Carlo method to obtain samples  for each weight based on a Gaussian distribution, whose mean is 0 and variance is equivalent to the device variance; Second, these samples will be added to the corresponding weights in the forward path in the training stage. Since only one Monte Carlo sample for each weight is required in each forward path, we can obtain the reasonable accuracy with the minor extra training time introduced by our proposed method.


Based on the proposed trainer, \todo{ptbNAS} is executed as follows.
The controller, trainer, and accuracy evaluator collaboratively search the parameters of neural architecture, quantization, and devices for higher accuracy while taking noises caused by hardware perturbation into account and proposing a variety of candidate architectures. This searching step includes four phases.
First, the controller predicts a quantized neural architecture and a type of device. 
Second, the identified architecture is trained by the trainer using the proposed weight perturbation aware training method. 
Third, the trained model is then evaluated by the accuracy evaluator to generate inference accuracy with noise. 
Finally, the accuracy will be the reward to update the controller for predicting new hyperparameters.

\raisebox{-1pt}{\large\ding{195}} \textbf{{Performance Evaluator.}}
Before entering the performance evaluator, we first conduct the circuit optimization.
We base the circuit optimization on NeuroSim \cite{chen2018neurosim}, and make modifications to support different quantization for network layers.
Based on the modified model, given a neural architecture $A$, a quantization $Q$, a device $D$, we can optimize the circuit and determine the parameters in \todo{circuit design $C$}.
Then, based on $C$ and the evaluation tool in \cite{peng2019dnn}, \todo{we can estimate the latency (Lat),  energy efficiency (Eng), and area (Area) for the implementation, which will be used in calculating the reward, as shown in Formula \ref{equ:reward}}.

Based on the above performance evaluator, the rNAS will fine tune quantization parameters of the candidate architectures to further integrate hardware metrics, including area, energy and latency into consideration. 
In the exploration, we will fix the neural architecture and device, so that there is no need to train the network from scratch to accelerate the search process.
Specifically, we open the switches $SA$ and $SD$, and close \todo{switches} $SQ$ and $SC$.
In each iteration, we will predict new quantization parameters for the identified neural architecture and device. 
Then, we will first obtain the inference accuracy via accuracy evaluator using the saved weights and the new quantization parameters.
Next, we will conduct the circuit optimization and obtain the hardware metrics including latency, energy, and area. 
Finally, we generate the reward in terms of the reward function, and update the controller based on the reward for the prediction in the next iteration.

\section{Experiments and Results}\label{sec:exp}

In this section, we will first present the experiment setup. Then the experimental results will be presented. 

\begin{table}[t]
  \centering
  \tabcolsep 1pt
  \renewcommand\arraystretch{1.5}
  \caption{\todoww{Experimental settings for three types of backbone on two datasets, CIFAR-10 and Nuclei.}}
    \begin{tabular}{|c|c|c|c|c|}
    \hline
    Spaces &  \# Layer & \# Filter & Filter H/W & FC Neuros \\
    \hline
    Res. Lim. &  8     & 24,36,48,64 & 1,3,5,7 & 64,128,256,512 \\
    VGG-Like Space &  11    & 128, 256,512,1024    & 1,3,5,7      &  256,512,1024,2048\\
    Enc-Dec-Like &  4,6,8,10 & 16,32,64,128 & 3     & - \\
    \hline
    \multicolumn{5}{l}{$\bullet$ Filter H/W: Height and width of filter; FC: Fully connection layer}
    \\
    \end{tabular}%
  \label{tab:exp_set}%
\end{table}%

\subsection{Experiment Setup}

\todoqw{In this work, we explore two machine learning tasks, image classification and object segmentation, to evaluate the proposed framework, NACIM. For the image classification task, similar to most existing works on CiM based neural accelerators \cite{patil2019mram,sun2018computing}, \todo{we} use the CIFAR-10 dataset \cite{krizhevsky2009learning};
while for the object segmentation, we apply the Nuclei dataset \cite{kumar2017dataset}
Table \ref{tab:exp_set} shows the neural architecture search spaces for these datasets. For CIFAR-10, we use a VGG-Like Space (VLS) backbone architecture, and an in-house constructed Resource Limited Space (RLS) backbone architecture. As to be shown in the results, the architectures in VLS require a large number of resources, which is not practical; and therefore, we introduced the RLS, which is designed for a resource limited scenario with sacrifices in accuracy. For Nuclei, the backbone architecture is encoder-decoder (Enc-Dec-Like, EDS), we explored different number of layers, and number of filters in each layer.}


\todoww{For the resource limited scenario (RLS), we also explore the Quantization space.} The quantization bit width of the activation and weight of each layer are searched separately. 
For each type of data, we determine the number of integer bits range from 0 to 3, and the number of fraction bits range from 0 to 6.





For the device and circuit, in this section, we use 4-bit ReRAM devices in the crossbar computation. The noise model of the device is from \cite{zhao2017rram}. We assume the current range of the device to be [0, 16 $uA$]. In each level of the device, the variation follows a Gaussian distribution, with a mean of 0 and standard deviation of 800$nA$. We assume the array size for crossbar to be $64 \times 64$. The updating rate of the controller is set to be 0.2 and the framework trains each candidate architecture for 30 epochs and searches for the optimal architecture for 500 episodes.
We pick the architectures with top 40 hardware noise aware inference accuracy from the searching results, and further fine-tune them with 200 training epochs for each network.



We search through layer-wise quantization parameters for each candidate architecture while assuming the underlying hardware to have the properties listed as follows: we use 4-bit ReRAMs as our CiM device and 16 level (4-bit) ADCs for the crossbar, chip clock frequency is 1 GHz, chip technology node is 32 nm. The memory voltage is 0.5 V and the chip voltage is 1.1 V. For each candidate architecture, the controller starts from the specifications provided by the previous search step, then performs 100 search steps to generate an optimized quantization scene for this architecture.

\subsection{\todoww{Exploration for Resource Limited Scenarios}} 

\begin{table}[t]
  \centering
  \tabcolsep 1pt
  \renewcommand\arraystretch{1.5}
  \caption{Comparison results between the proposed approches and the state-of-the-art QuantNAS without the consideration of the device durding the search process.}
    \begin{tabular}{|c|c|c|c|c|c|c|}
    \hline
    \multirow{2}{*}{Approach} & \multirow{2}{*}{Accuracy} & Acc w/  & Area   & EDP  & Speed & E.-E. \\
    & & variation & ($\mu m^2$) & ($pJ*ns$) & (TOPs) & (TOPs/W)\\
    \hline
    QuantNAS &  84.92\%     &   8.48\%    &   $3.24*10^{6}$  & $ 8.08*10^{12}$ & 0.285 & 5.14\\
    \hline
    \todo{ptbNAS} &    \todo{74.28\%}   &  \todo{72.18\%}     &   \todo{$2.57*10^{6}$}  & \todo{$7.9*10^{12}$} & \todo{0.117} &\todo{4.99}\\
    \hline
    \placeholder$_{hw}$ &   73.58\%    &   70.12\%    &   $\bm{1.78*10^{6}}$    & $\bm{2.21*10^{12}}$ & 0.204 & 12.3 \\
    \hline
    \placeholder$_{sw}$ &   73.88\%    &   \textbf{73.45\%}    &   $1.97*10^{6}$    & $3.76*10^{12}$ & 0.234 & \textbf{16.3} \\
    \hline
    \end{tabular}%
  \label{tab:results}%
\end{table}%

\todoww{In this subsection, we report the exploration results of employing the resource limited search space (RLS) for CIFAR-10 dataset.
We first compare the proposed NACIM to the existing approach; then, we demonstrate design space exploration results with the tradeoffs in terms of multiple metrics.}

\vspace{3pt}
\noindent\textit{(1) Comparison Results to State-of-the-Art NAS}
\vspace{3pt}

First, we show the exploration results of different searching methods in Table~\ref{tab:results}. 
``QuantNAS'' indicates the state-of-the-art \todo{quantization-architecture} co-exploration method proposed in \cite{Lu2019Neural}, \todoww{where the standard training procedure is conducted}. 
``\todo{ptbNAS}'' indicates the noise-aware training and searching method proposed in this work, where the switch combination is set as $SA=1,SQ=1,SD=1,SC=0$.
Kindly note that the QuantNAS is the basis of ptbNAS, but ptbNAS integrate the noise-awareness during the search process.
``\placeholder'' indicates the noise-aware training and searching method along with the hardware resource-aware quantization search, which combines $\todo{ptbNAS}$ and $rNAS$.
Please note that ``\placeholder'' can obtain a serials of solutions on Pareto frontier.
We use notation ``\placeholder$_{hw}$'' and ``\placeholder$_{sw}$'' to represent the solution with maximum hardware efficiency and that with maximum accuracy, respectively.
\todoww{The detailed architectures identified by these two approaches are summarized in Table \ref{tab:nas_arch}.} 
For comparison, we obtain the accuracy of all architectures without noise, as shown in column ``Accuracy''.
We then compare the accuracy after considering the device variation in column ``Acc w/ variation''.
We employ the same circuit optimization procedure, and obtain the hardware efficiency metrics, including area and energy delay product (EDP), speed (TOPs), and energy efficiency (TOPs/W).

\begin{table}[t]
  \centering
  \tabcolsep 11pt
  \renewcommand\arraystretch{1.2}
  \caption{\todoww{Identified neural architecture with quantization information for $NACIM_{hw}$ AND $NACIM_{sw}$.}}
    \begin{tabular}{|c|c|c|}
    \hline
    \multicolumn{1}{|c|}{Layer} & \multicolumn{1}{c|}{$NACIM_{hw}$} & \multicolumn{1}{c|}{$NACIM_{sw}$} \\
    \hline
    conv1 & (3, 5, 64, 0, 2, 6, 2, 6) & (5, 5, 64, 0, 1, 5, 3, 6) \\
    conv2 & (3, 1, 48, 0, 1, 2, 1, 2) & (3, 1, 48, 0, 3, 2, 1, 6) \\
    conv3 & (1, 3, 48, 1, 2, 6, 1, 3) & (1, 3, 48, 1, 2, 0, 3, 5) \\
    conv4 & (5, 3, 64, 1, 1, 2, 0, 4) & (5, 5, 64, 1, 2, 0, 0, 4) \\
    conv5 & (1, 1, 64, 1, 0, 1, 1, 3) & (1, 1, 64, 1, 1, 4, 2, 3) \\
    conv6 & (3, 3, 24, 0, 1, 1, 2, 5) & (3, 3, 24, 0, 0, 1, 0, 5) \\
    fc1   & (256, -, -, -, 3, 5, 1, 3) & (256, - , -, 1, 2, 2, 3, 6) \\
    fc2   & (64, -, -, -, 1, 3, 2, 6) & (64, -, -, 0, 0, 2, 0, 2) \\
    \hline
    \multicolumn{3}{l}{Parameters are (FH,FW,\#F,P,WQ\_int,WQ\_frac,AQ\_int,AQ\_frac)} \\
    \multicolumn{3}{l}{$\bullet$ FH/FW: Filter Height/Width; \#F: Num of Filter; P: Pooling or not}\\
    \multicolumn{3}{l}{$\bullet$ xQ\_int: \# bits in weight (x=W) or activation (x=A) for integer} \\
    \multicolumn{3}{l}{$\bullet$ xQ\_frac: \# bits in weight (x=W) or activation (x=A) for fraction} \\
    \end{tabular}%
  \label{tab:nas_arch}%
\end{table}%

Results in Table~\ref{tab:results} shows that \todo{QuantNAS} can find architecture with the highest accuracy. However, when it is employed for computing-in-memory circuit with variation, it has a drastic accuracy loss from 84.92\% to 8.48\%, rendering the architecture to be useless.
On the contrary, with consideration of device variation in training process, the network accuracies of \todo{ptbNAS},  \placeholder$_{hw}$,  \placeholder$_{sw}$ on computing-in-memory circuit are 72.18\%, 70.12\%, 73.45\%, respectively.
What is more, the accuracy loss for \placeholder$_{sw}$ is only 0.43\%.

We can also observe from the table that by employing the cross-layer optimization, \placeholder$_{hw}$ can obtain the best hardware efficiency.
Compared with QuantNAS, \placeholder$_{hw}$ achieves 1.82$\times$ reduction on area and 3.66$\times$ improvement on energy delay product.
Compared with \todo{ptbNAS}, these figures are 14.01\% and 1.89$\times$, respectively.
Compared with \placeholder$_{sw}$, these figures are 9.64\% and 1.70$\times$, respectively.
These results demonstrate the capability of \placeholder~to synthesize the cost-effective computing-in-memory chips.

\begin{figure*}[t]
  \centering
  \includegraphics[width=0.99\linewidth]{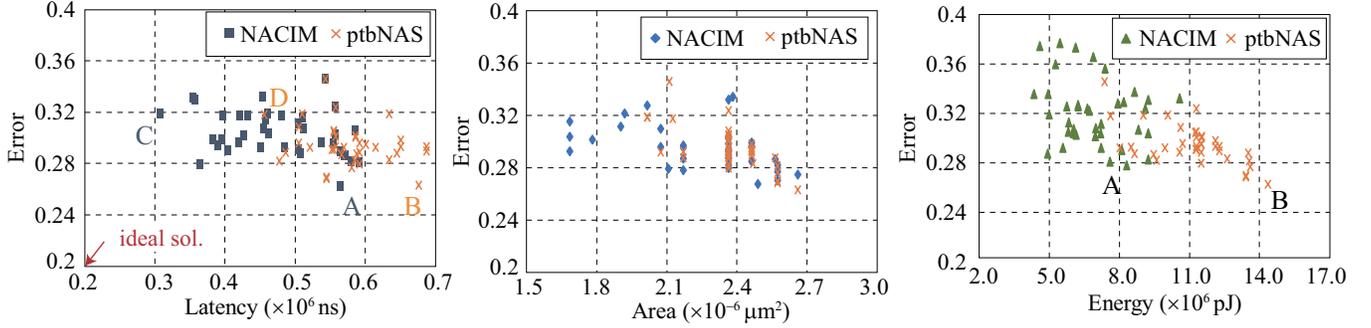}
  \vspace{-10pt}
  \caption{\todoww{Inference results by applying bi-objective optimizations: (left) accuracy vs. latency; (middle) accuracy vs. chip area; (right) accuracy vs. energy.}}
  \label{fig:error-latency}
  \label{fig:error-area}
  \label{fig:error-energy}
\end{figure*}

Another observation is that the architectures identified by both \todo{QuantNAS} and \placeholder$_{sw}$ achieve slightly higher speed than that by \placeholder$_{hw}$. 
This is because \placeholder$_{hw}$ finds many simple structures with fewer operations, but the latency is not improved accordingly since other designs can have more processing elements.
In the comparison of energy efficiency, 
\placeholder$_{hw}$ achieves $2.39 \times$ higher energy efficiency than \todo{QuantNAS}.
\placeholder$_{sw}$ achieves $3.17\times$ higher energy efficiency, reaching up to 16.3 TOPs/W.
The above observations clearly show the importance of conducting cross-layer optimization to obtain useful neural architectures for hardware efficient computing-in-memory architecture.




\vspace{3pt}
\noindent\textit{(2) Results of Bi-Objective Optimization}
\vspace{3pt}

\begin{figure}[t]
  \centering
  \includegraphics[width=0.8\linewidth]{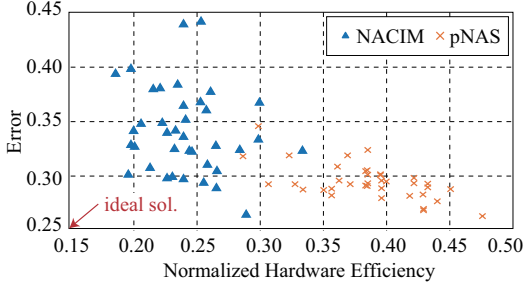}
  \vspace{-10pt}
  \caption{Multi-objective optimization: inference error vs. normalized hardware efficiency. The hardware efficiency is the weighted sum of hardware area, energy and latency.}
  \label{fig:error-punishment}
\end{figure} 

Next, we report the design space exploration results of both \todo{ptbNAS} and NACIM with bi-objective optimization: maximizing the accuracy and hardware performance.
Here, the accuracy is obtained by executing the neural network on computing-in-memory chip with variation.
And we carry out three sets of experiments to optimize each hardware performance metric, including latency, area, and energy, separately.
The reward function is calculated based on these metrics, \todo{as shown} in Formula \ref{framework}, where we set $\beta$ to be 0.5 to co-optimize network accuracy and hardware efficiency.
In the bi-objective optimization, function $f$ will only return the value of one metric, and we will extend to multi-objective optimization in the next subsection.

Figure \ref{fig:error-latency} shows the design space exploration in terms of accuracy and latency.
In this figure, the x-aixs and y-aixs represent the latency and error, respectively.
Each rectangle stands for a design identified by \placeholder~and each cross stands for a design identified by \todo{ptbNAS}.
For all multi-objective results, the ideal solutions will be on the bottom-left corner, as shown in this figure.

From the results, we can see that by considering the cross-layer optimization, \placeholder~can significantly push forward the Pareto frontier between accuracy and latency.
This is because \placeholder~will generate the reward using the weighted accuracy and latency, which can improve the latency by find better circuit design and guarantee accuracy at the same time.
Specifically, for the comparison between solutions with the highest accuracy (design A for \placeholder~, and B for \todo{ptbNAS}), we can see that A's accuracy (73.77\%) is higher than B's accuracy (73.69\%). What is more, design A reduces latency by 16.63\%.
For the comparison between solutions with the lowest latency, we can see that \placeholder~(design C) achieves the same accuracy but 32.49\% lower latency, compared with \todo{ptbNAS} (design D).





We further conduct experiments on optimizing area and energy. We observed similar results.
The results are shown in Figures \ref{fig:error-area} and \ref{fig:error-energy}. 
There is one interesting observation in exploring the design space for accuracy and energy tradeoffs, which is shown in Figure \ref{fig:error-energy}. The figure shows that \todo{ptbNAS} can find solutions with higher accuracy against the NACIM.
For example, in the figure, design A identified by \placeholder~has 1\% accuracy loss against design B, which is identified by \todo{ptbNAS}. However, \placeholder~achieves 1.73$\times$ higher energy efficiency.
Here, both designs have the same neural architecture but different quantization.
In order to obtain high energy efficiency, \placeholder~employs lower bit-width precision.
We can avoid such accuracy loss by increasing the scaling variable $\beta$ in the reward function in Formula \ref{framework}.



All above observations verify the importance of conducting bi-objective optimization instead of mono-objective optimization on accuracy.



\begin{figure}[t]
  \centering
  \includegraphics[width=1\linewidth]{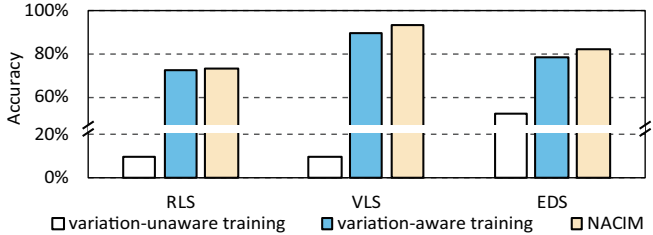}
  \vspace{-20pt}
  \caption{\todoww{On RLS (Resource Limited), VLS (VGG-like), EDS (Encoder-decoder-like) search spaces, the comparison results in accuracy obtained by three approaches, where variation-unaware and variation-aware training are based on a same fixed architecture; NACIM opens architecture search space.}}
  \label{fig:NACIMAdvance}
\end{figure}

\vspace{3pt}
\noindent\textit{(3) Results of Multi-Objective Optimization}
\vspace{3pt}

Figure \ref{fig:error-punishment} shows the design space exploration tradeoffs between accuracy and the normalized hardware efficiency.
The normalized hardware efficiency is calculated based on weighted hardware metrics, including latency, area, and energy, which is represented by the x-axis. Each hardware component has a same weight and the total normalized hardware efficiency has the consists of half of the reward and inference accuracy takes another half.
An interesting observation from the results is that compared with the bi-objective optimization, \todo{\placeholder~found} more architectures with lower accuracy.
This is because the weights for accuracy in calculating the reward is decreased.
However, we can still can find the solution with the highest accuracy, and achieves 1.65$\times$ improvement on hardware efficiency.


\subsection{\todoww{Scalability of NACIM}}

\todoqw{The previous subsection has shown the advantages of NACIM over the existing techniques. In this subsection, we further evaluate the scalability of NACIM on (1) a larger backbone architecture on CIFAR-10; (2) a more complicated machine learning task, object segmentation.}

\begin{table}[t]
  \centering
  \tabcolsep 5pt
  \renewcommand\arraystretch{1.2}
  \caption{\todoww{Comparison results of accuracy and hardware usage of architectures from Figure \ref{fig:NACIMAdvance} and their CiM implementations.}}
    \begin{tabular}{|c|c|c|c|c|c|c|}
    \hline
    \multirow{2}{*}{Search Space} & \multirow{2}{*}{Approach} & Accuracy  & Area   & EDP  \\
     & & or IOU& ($\mu m^2$) & ($pJ*ns$)\\
    \hline
    \multirow{2}{*}{RLS} & Baseline &  72.18\%    &   $2.57*10^{6}$  & $ 7.9*10^{12}$ \\
    \cline{2-5}
    & \textbf{\placeholder} &   \textbf{73.45\%}   &   $\bm{1.97*10^{6}}$    & $\bm{3.76*10^{12}}$  \\
    
    \hline
    \multirow{2}{*}{VLS} & Baseline &  90.06\%    &   $4.57*10^{8}$  & $ 1.42*10^{15}$ \\
    \cline{2-5}
    & \textbf{\placeholder} &   \textbf{93.12\%}   &   $\bm{4.75*10^{8}}$    & $\bm{5.54*10^{15}}$  \\
    
    \hline
    \multirow{2}{*}{EDS} & Baseline &  0.788    &   $7.88*10^{7}$  & $ 3.86*10^{14}$ \\
    \cline{2-5}
    & \textbf{\placeholder} &   \textbf{0.824}  &   $\bf{4.05*10^{7}}$    & $\bf{1.96*10^{14}}$  \\
    \hline
    \end{tabular}%
  \label{tab:ACC_HW}%
\end{table}%

\todoqw{Figure \ref{fig:NACIMAdvance} demonstrates the results of accuracy comparison among (1) variation-unaware training, (2) variation-aware training, (3) NACIM for three different backbone architectures in terms of the search space, where inference is conducted on a CiM system with non-negligible devices variation. Note that the first two methods are based on a fixed neural architecture while NACIM explores different neural architectures; specifically, RLS uses an architecture explored by ptbNAS, VLS is based on the original VGG-11 architecture, and EDS employs 4 layers encoder-decoder.}


\todoqw{Our experimental results clearly show that if device variation is not considered during training, inference accuracy will be unacceptable; for RLS, VLS and EDS, the accuracy results are 9.8\%, 9.6\%, 0.525, respectively. Another observation is that the proposed variation-aware training can significantly improve the accuracy. In addition, after we enlarge the architecture search space, we can identify neural architectures with better accuracy (details can be found in Table IV).}


\todoqw{Finally, we report the accuracy and hardware trade-off in Table \ref{tab:ACC_HW}, where Baseline indicates the solution that applies the fixed architecture and the proposed variation-aware training procedure. Results from this table clearly show that taking VGG-11 in VLS as a backbone leads to excessive (i.e., larger than 100 times) area and energy-delay-product (EDP), compared with the solution generated in RLS. Second, the proposed NACIM framework can be applied and be effect in different backbones. Specifically, with a larger backbone, like VLS, the proposed NACIM can achieve up to 93.13\% accuracy. But it consumes much more hardware compared with a smaller backbone, like that in RLS.}

\section{conclusion and Future Work}\label{sec:conclusion}
In this work, we formally defined cross-layer optimization problem for automatically identifying neural architectures on computing-in-memory (CiM) platform.
We devised a novel neural architecture search framework that gives flexibility for designers to set different optimization goal.
We further integrate a trainer with the consideration of device variation in our framework.
In experiments, we first demonstrated the importance of finding a robust neural architecture in terms of the device variation in CiM, which may lead the neural architectures that apply the existing NAS to be useless due to dramatic accuracy loss.
We further showed that the cross-layer optimization can identify the robust neural architecture with 0.45\% accuracy loss after considering variation, and maximize hardware efficiency to achieve 16.3 TOPs/W energy efficiency.

\todoqw{Our experimental results have demonstrated the effectiveness of the hardware perturbation aware training procedure.  As future work, we will investigate how to optimize the trainer to speed up the training procedure and improve accuracy. One potential way for speedup is to replace the Monte Carlo sampling method by the Quasi-Monte Carlo method.
}

\bibliography{example_paper}

\begin{thebibliography}{10}
\providecommand{\url}[1]{#1}
\csname url@samestyle\endcsname
\providecommand{\newblock}{\relax}
\providecommand{\bibinfo}[2]{#2}
\providecommand{\BIBentrySTDinterwordspacing}{\spaceskip=0pt\relax}
\providecommand{\BIBentryALTinterwordstretchfactor}{4}
\providecommand{\BIBentryALTinterwordspacing}{\spaceskip=\fontdimen2\font plus
\BIBentryALTinterwordstretchfactor\fontdimen3\font minus
  \fontdimen4\font\relax}
\providecommand{\BIBforeignlanguage}[2]{{%
\expandafter\ifx\csname l@#1\endcsname\relax
\typeout{** WARNING: IEEEtran.bst: No hyphenation pattern has been}%
\typeout{** loaded for the language `#1'. Using the pattern for}%
\typeout{** the default language instead.}%
\else
\language=\csname l@#1\endcsname
\fi
#2}}
\providecommand{\BIBdecl}{\relax}
\BIBdecl

\bibitem{krizhevsky2012imagenet}
A.~Krizhevsky \emph{et~al.}, ``Imagenet classification with deep convolutional
  neural networks,'' in \emph{Proc. of NIPS}, 2012, pp. 1097--1105.

\bibitem{redmon2018yolov3}
J.~Redmon and A.~Farhadi, ``Yolov3: An incremental improvement,'' \emph{arXiv
  preprint arXiv:1804.02767}, 2018.

\bibitem{gao2018object}
H.~Gao, B.~Cheng, J.~Wang, K.~Li, J.~Zhao, and D.~Li, ``Object classification
  using cnn-based fusion of vision and lidar in autonomous vehicle
  environment,'' \emph{IEEE Transactions on Industrial Informatics}, vol.~14,
  no.~9, pp. 4224--4231, 2018.

\bibitem{xu2019whole}
X.~Xu, T.~Wang, Y.~Shi, H.~Yuan, Q.~Jia, M.~Huang, and J.~Zhuang, ``Whole heart
  and great vessel segmentation in congenital heart disease using deep neural
  networks and graph matching,'' in \emph{International Conference on Medical
  Image Computing and Computer-Assisted Intervention}.\hskip 1em plus 0.5em
  minus 0.4em\relax Springer, 2019, pp. 477--485.

\bibitem{zhang2018thundervolt}
J.~Zhang, K.~Rangineni, Z.~Ghodsi, and S.~Garg, ``Thundervolt: enabling
  aggressive voltage underscaling and timing error resilience for energy
  efficient deep learning accelerators,'' in \emph{Proceedings of the 55th
  Annual Design Automation Conference}.\hskip 1em plus 0.5em minus 0.4em\relax
  ACM, 2018, p.~19.

\bibitem{zhang2018fate}
J.~J. Zhang and S.~Garg, ``Fate: fast and accurate timing error prediction
  framework for low power dnn accelerator design,'' in \emph{Proceedings of the
  International Conference on Computer-Aided Design}.\hskip 1em plus 0.5em
  minus 0.4em\relax ACM, 2018, p.~24.

\bibitem{jiang2019achieving}
W.~Jiang, E.~H.-M. Sha, X.~Zhang, L.~Yang, Q.~Zhuge, Y.~Shi, and J.~Hu,
  ``Achieving super-linear speedup across multi-fpga for real-time dnn
  inference,'' \emph{ACM Transactions on Embedded Computing Systems (TECS)},
  vol.~18, no.~5s, p.~67, 2019.

\bibitem{xu2018resource}
X.~Xu \emph{et~al.}, ``Resource constrained cellular neural networks for
  real-time obstacle detection using fpgas,'' in \emph{Proc. of ISQED}.\hskip
  1em plus 0.5em minus 0.4em\relax IEEE, 2018, pp. 437--440.

\bibitem{jiang2018heterogeneous}
W.~Jiang \emph{et~al.}, ``Heterogeneous fpga-based cost-optimal design for
  timing-constrained cnns,'' \emph{IEEE TCAD}, 2018.

\bibitem{zoph2016neural}
B.~Zoph and Q.~V. Le, ``Neural architecture search with reinforcement
  learning,'' in \emph{International Conference on Learning Representations
  (ICLR)}, 2017.

\bibitem{zoph2017learning}
B.~Zoph, V.~Vasudevan, J.~Shlens, and Q.~V. Le, ``Learning transferable
  architectures for scalable image recognition,'' in \emph{IEEE conference on
  Computer Vision and Pattern Recognition (CVPR)}, 2018, pp. 8697--8710.

\bibitem{real2017large}
E.~Real, S.~Moore, A.~Selle, S.~Saxena, Y.~L. Suematsu, J.~Tan, Q.~Le, and
  A.~Kurakin, ``Large-scale evolution of image classifiers,'' \emph{arXiv
  preprint arXiv:1703.01041}, 2017.

\bibitem{liu2017hierarchical}
H.~Liu, K.~Simonyan, O.~Vinyals, C.~Fernando, and K.~Kavukcuoglu,
  ``Hierarchical representations for efficient architecture search,''
  \emph{arXiv preprint arXiv:1711.00436}, 2017.

\bibitem{nekrasov2019architecture}
V.~Nekrasov, H.~Chen, C.~Shen, and I.~Reid, ``{Architecture Search of Dynamic
  Cells for Semantic Video Segmentation},'' \emph{arXiv preprint
  arXiv:1904.02371}, 2019.

\bibitem{liu2018darts}
H.~Liu, K.~Simonyan, and Y.~Yang, ``Darts: Differentiable architecture
  search,'' \emph{arXiv preprint arXiv:1806.09055}, 2018.

\bibitem{tan2018mnasnet}
M.~Tan, B.~Chen, R.~Pang, V.~Vasudevan, and Q.~V. Le, ``Mnasnet: Platform-aware
  neural architecture search for mobile,'' \emph{arXiv preprint
  arXiv:1807.11626}, 2018.

\bibitem{cai2018efficient}
H.~Cai, T.~Chen, W.~Zhang, Y.~Yu, and J.~Wang, ``Efficient architecture search
  by network transformation.''\hskip 1em plus 0.5em minus 0.4em\relax AAAI,
  2018.

\bibitem{wu2018fbnet}
B.~Wu, X.~Dai, P.~Zhang, Y.~Wang, F.~Sun, Y.~Wu, Y.~Tian, P.~Vajda, Y.~Jia, and
  K.~Keutzer, ``Fbnet: Hardware-aware efficient convnet design via
  differentiable neural architecture search,'' \emph{arXiv preprint
  arXiv:1812.03443}, 2018.

\bibitem{cai2018proxylessnas}
H.~Cai, L.~Zhu, and S.~Han, ``\todo{Proxylessnas: Direct neural architecture
  search on target task and hardware},'' \emph{arXiv preprint
  arXiv:1812.00332}, 2018.

\bibitem{dai2019chamnet}
X.~Dai, P.~Zhang, B.~Wu, H.~Yin, F.~Sun, Y.~Wang, M.~Dukhan, Y.~Hu, Y.~Wu,
  Y.~Jia \emph{et~al.}, ``\todo{Chamnet: Towards efficient network design
  through platform-aware model adaptation},'' in \emph{Proceedings of the IEEE
  Conference on Computer Vision and Pattern Recognition}, 2019, pp.
  11\,398--11\,407.

\bibitem{stamoulis2019single}
D.~Stamoulis, R.~Ding, D.~Wang, D.~Lymberopoulos, B.~Priyantha, J.~Liu, and
  D.~Marculescu, ``\todo{Single-path nas: Designing hardware-efficient convnets
  in less than 4 hours},'' \emph{arXiv preprint arXiv:1904.02877}, 2019.

\bibitem{jiang2019accuracy}
W.~Jiang, X.~Zhang, E.~H.-M. Sha, L.~Yang, Q.~Zhuge, Y.~Shi, and J.~Hu,
  ``Accuracy vs. efficiency: Achieving both through fpga-implementation aware
  neural architecture search,'' in \emph{Proceedings of the 56th Annual Design
  Automation Conference 2019}.\hskip 1em plus 0.5em minus 0.4em\relax ACM,
  2019, p.~5.

\bibitem{hao2019fpga}
C.~Hao, X.~Zhang, Y.~Li, S.~Huang, J.~Xiong, K.~Rupnow, W.-m. Hwu, and D.~Chen,
  ``\todo{FPGA/DNN Co-Design: An Efficient Design Methodology for 1oT
  Intelligence on the Edge},'' in \emph{2019 56th ACM/IEEE Design Automation
  Conference (DAC)}.\hskip 1em plus 0.5em minus 0.4em\relax IEEE, 2019, pp.
  1--6.

\bibitem{jiang2019hardware}
W.~Jiang, L.~Yang, E.~Sha, Q.~Zhuge, S.~Gu, Y.~Shi, and J.~Hu,
  ``Hardware/software co-exploration of neural architectures,'' \emph{arXiv
  preprint arXiv:1907.04650}, 2019.

\bibitem{wong2018inmemory}
D.~Ielmini and H.-S.~P. Wong, ``In-memory computing with resistive switching
  devices,'' in \emph{Nature Electronics}.\hskip 1em plus 0.5em minus
  0.4em\relax Nature, 2018, p. 333.

\bibitem{sze2017tutorial}
Y.-H. C. T.-J.~Y. Sze, Vivienne and J.~S. Emer, ``Efficient processing of deep
  neural networks: A tutorial and survey,'' in \emph{Proceedings of the
  IEEE}.\hskip 1em plus 0.5em minus 0.4em\relax IEEE, 2017, pp. 2295--2329.

\bibitem{shafiee2016isaac}
A.~N. N. M.-R. B. J. P. S. M. H. R. S.~W. Shafiee, Ali and V.~Srikumar,
  ``Isaac: A convolutional neural network accelerator with in-situ analog
  arithmetic in crossbars,'' \emph{ACM SIGARCH Computer Architecture News}, pp.
  14--26, 2016.

\bibitem{biswas2018convram}
A.~Biswas and A.~P. Chandrakasan, ``Conv-ram: An energy-efficient sram with
  embedded convolution computation for low-power cnn-based machine learning
  applications,'' in \emph{2018 IEEE International Solid-State Circuits
  Conference-(ISSCC)}.\hskip 1em plus 0.5em minus 0.4em\relax IEEE, 2018, pp.
  488--490.

\bibitem{kang2018in_memory}
S.~L. S.~G. Kang, Mingu and N.~Shanbhag, ``An in-memory vlsi architecture for
  convolutional neural networks,'' in \emph{IEEE Journal on Emerging and
  Selected Topics in Circuits and Systems}, 2018, pp. 494--505.

\bibitem{chi2016prime}
S.~L. C. X.-T. Z. J. Z. Y. L. Y.~W. Chi, Ping and Y.~Xie, ``Prime: A novel
  processing-in-memory architecture for neural network computation in
  reram-based main memory,'' \emph{In ACM SIGARCH Computer Architecture News,
  vol. 44, no. 3, pp. 27-39. IEEE Press}, pp. 27--39, 2016.

\bibitem{xu2018scaling}
X.~Xu \emph{et~al.}, ``Scaling for edge inference of deep neural networks,''
  \emph{Nature Electronics}, vol.~1, no.~4, p. 216, 2018.

\bibitem{xu2018quantization}
X.~Xu, Q.~Lu, L.~Yang, S.~Hu, D.~Chen, Y.~Hu, and Y.~Shi, ``Quantization of
  fully convolutional networks for accurate biomedical image segmentation,'' in
  \emph{Proceedings of the IEEE Conference on Computer Vision and Pattern
  Recognition}, 2018, pp. 8300--8308.

\bibitem{zhang2019compact}
J.~J. Zhang, P.~Raj, S.~Zarar, A.~Ambardekar, and S.~Garg, ``Compact: On-chip
  compression of activations for low power systolic array based cnn
  acceleration,'' \emph{ACM Transactions on Embedded Computing Systems (TECS)},
  vol.~18, no.~5s, p.~47, 2019.

\bibitem{zhao2017rram}
H.~W. B. G.-Q. Z. W. W. S. W. Y.~X. Zhao, Meiran, ``Investigation of
  statistical retention of filamentary analog rram for neuromophic computing,''
  \emph{IEEE International Electron Devices Meeting (IEDM)}, pp. 39--4, 2017.

\bibitem{liu2019auto}
C.~Liu \emph{et~al.}, ``Auto-deeplab: Hierarchical neural architecture search
  for semantic image segmentation,'' in \emph{Proc. of CVPR}, 2019, pp. 82--92.

\bibitem{peng2019video}
W.~Peng \emph{et~al.}, ``Video action recognition via neural architecture
  searching,'' in \emph{Proc. of ICIP}.\hskip 1em plus 0.5em minus 0.4em\relax
  IEEE, 2019, pp. 11--15.

\bibitem{ben2018error}
S.~W. Feinberg, Ben and E.~Ipek, ``Maing memristive neural network accelerators
  reliable,'' \emph{IEEE International Symposium on High Performance Computer
  Architecture (HPCA)}, pp. 52--65, 2018.

\bibitem{peng2019dnn}
X.~Peng \emph{et~al.}, ``Dnn+neurosim: An end-to-end benchmarking framework for
  compute-in-memory accelerators with versatile device technologies,'' in
  \emph{Proc. of IEDM}, 2019.

\bibitem{chen2018neurosim}
P.~{Chen}, X.~{Peng}, and S.~{Yu}, ``Neurosim: A circuit-level macro model for
  benchmarking neuro-inspired architectures in online learning,'' \emph{IEEE
  Transactions on Computer-Aided Design of Integrated Circuits and Systems},
  vol.~37, no.~12, pp. 3067--3080, Dec 2018.

\bibitem{williams1992simple}
R.~J. Williams, ``Simple statistical gradient-following algorithms for
  connectionist reinforcement learning,'' \emph{Machine learning}, vol.~8, no.
  3-4, pp. 229--256, 1992.

\bibitem{Lu2019Neural}
Q.~Lu, W.~Jiang, X.~Xu, Y.~Shi, and J.~Hu, ``On neural architecture search for
  resource-constrained hardware platforms,'' in \emph{International Conference
  on Computer-Aided Design (ICCAD)}.\hskip 1em plus 0.5em minus 0.4em\relax
  ACM, 2019, p.~1.

\bibitem{zhang2019resilience}
L.-Y.~C. Zhang, Bonan and N.~Verma, ``Stochastic data-driven hardware
  resilience to efficiently train inference models for stochastic hardware
  implementations,'' in \emph{ICASSP 2019-2019 IEEE International Conference on
  Acoustics, Speech and Signal Processing (ICASSP)}, 2019.

\bibitem{patil2019mram}
A.~D. Patil, H.~Hua, S.~Gonugondla, M.~Kang, and N.~R. Shanbhag, ``An
  mram-based deep in-memory architecture for deep neural networks,'' in
  \emph{2019 IEEE International Symposium on Circuits and Systems
  (ISCAS)}.\hskip 1em plus 0.5em minus 0.4em\relax IEEE, 2019, pp. 1--5.

\bibitem{sun2018computing}
X.~Sun, R.~Liu, X.~Peng, and S.~Yu, ``Computing-in-memory with sram and rram
  for binary neural networks,'' in \emph{2018 14th IEEE International
  Conference on Solid-State and Integrated Circuit Technology (ICSICT)}.\hskip
  1em plus 0.5em minus 0.4em\relax IEEE, 2018, pp. 1--4.

\bibitem{krizhevsky2009learning}
A.~Krizhevsky, G.~Hinton \emph{et~al.}, ``\todo{Learning multiple layers of
  features from tiny images},'' 2009.

\bibitem{kumar2017dataset}
N.~Kumar, R.~Verma, S.~Sharma, S.~Bhargava, A.~Vahadane, and A.~Sethi,
  ``\todo{A dataset and a technique for generalized nuclear segmentation for
  computational pathology},'' \emph{IEEE transactions on medical imaging},
  vol.~36, no.~7, pp. 1550--1560, 2017.

\end{thebibliography}
\bibliographystyle{IEEEtran}

\end{document}